\documentclass[11pt]{article}
\usepackage{mtsummit2019}
\usepackage{times}
\usepackage{url}
\usepackage{latexsym}
\usepackage[small,bf]{caption} 
\usepackage{multirow}
\usepackage[utf8]{inputenc}
\usepackage{makecell} 
\usepackage{todonotes}

\usepackage{footnote}
\makesavenoteenv{tabular}
\makesavenoteenv{table}

\title{Post-editese: an Exacerbated Translationese}

\author{Antonio Toral\\
Center for Language and Cognition\\
University of Groningen\\
  The Netherlands\\  {\tt a.toral.ruiz@rug.nl}}

\date{}

\begin{document}
\maketitle
\begin{abstract}
Post-editing (PE) machine translation (MT) is widely used for dissemination because it leads to higher productivity than human translation from scratch (HT).
In addition, PE translations are found to be of equal or better quality than HTs. However, most such studies measure quality solely as the number of errors.
We conduct a set of computational analyses in which we compare PE against HT on three different datasets that cover five translation directions with measures that address different translation universals and laws of translation: simplification, normalisation and interference.
We find out that PEs are simpler and more normalised and have a higher degree of interference from the source language than HTs.

\end{abstract}

\section{Introduction}

Machine translation (MT) is nowadays widely used in industry for dissemination purposes by means of post-editing (PE, also referred to as PEMT in the literature), a machine-assisted approach to translation that results in notable increases in translation productivity compared to unaided human translation (HT), as shown in numerous research studies, e.g. Plitt and Masselot~\shortcite{plitt2010productivity}.

In theory, one would claim that HTs\footnote{By HT we refer to translations produced by a human from scratch, i.e. without the assistance of an MT system or any other computer-assisted technology, e.g. translation memories.} and PE translations 
are clearly different, since, in the translation workflow of the latter, the translator is primed by the output of an MT system~\cite{Green2013}, resulting in a translation that should then contain, to some extent, the footprint of that MT system. Because of this, one would conclude that HT should be preferred over PE, as the former should be more natural and adhere more closely to the norms of the target language. However, many research studies have shown that the quality of PE is comparable to that of HT 
or even better, e.g. Koponen~\shortcite{Koponen2016}, and, according to one study~\cite{Bowker2015}, native speakers do not have a clear preference for HT over PE. 

In this paper we conduct a set of computational analyses on several datasets that contain HTs and PEs, involving different language directions and domains as well as PE performed according to different guidelines (e.g. full versus light). Our aim is to find out whether HT and PE differ significantly in terms of different phenomena. 
Since previous research has proven the existence of translationese, i.e. the fact that HT and original text exhibit different characteristics, our current research can be framed as a quest to find out whether there is evidence of {\it post-editese}, i.e. the fact that HT (or translationese) and PE would be different.

The characteristics of translationese can be grouped along the so-called universal features of translation or translation universals~\cite{baker1993corpus}, 
namely simplification, normalisation (also referred to as homogeneisation) and explicitation. In addition to these three, interference is recognised as a fundamental law of translation~\cite{toury2012descriptive}: 
``phenomena pertaining to the make-up of the source text tend to be transferred to the target text". In a nutshell, compared to original texts, translations tend to be simpler, more standardised, and more explicit and they retain some characteristics that pertain to the source language.

In this study then we study the existence of post-editese by conducting a set of computational analyses that fall into three\footnote{Explicitation is not addressed in our experiments.} out of these four categories. 
With these analyses we aim to answer a number of research questions:
\begin{itemize}
\item \textbf{RQ1}. Does post-editese exist? I.e. is there evidence that PE exhibits different characteristics than HT?
\item \textbf{RQ2}. If the answer to RQ1 is yes, then which are the main characteristics of PE? I.e. how does it differ from HT?
\item \textbf{RQ3}. If the answer to RQ1 is yes, then, are there different post-editeses? I.e. are there any characteristics that distinguish the post-editese produced by MT systems that follow different paradigms (rule-based, statistical phrase-based and neural)?
\end{itemize}


The rest of this paper is organised as follows.
Section~\ref{s_related_work} provides an overview of the related work.
Section~\ref{s_experiments} covers the experimental setup and the experiments conducted.
Finally, Section~\ref{s_conclusions} presents our conclusions and suggests lines of future work.

\section{Related Work}\label{s_related_work}

Many research studies carried out during the last decade have compared the quality of HT and PE. These have shown that the quality of PE is comparable to that of HT, e.g. García~\shortcite{garcia2010}, or even better, e.g. Guerberof~\shortcite{guerberof2009productivity} and Plitt and Masselot~\shortcite{plitt2010productivity}.
In these studies quality is typically measured in terms of the number of mistakes in each translation condition.
However, they do not take into account other relevant aspects that may flag important differences between HTs and PEs, such as the perspective of the end-user or the presence of different phenomena in both types of translations.
A recent strand of work targets precisely these, (i) by collecting the preference of end users between HT and PE and (ii) by analysing the characteristics of both types of translations.
In the following subsections we report on recent work conducted in both of these two research lines. 

\subsection{Preference between HT and PE}

Fiederer and O'Brien~\shortcite{fiederer2009quality} compared HT and PE for the English-to-German translation of text from a software user manual.
Participants ranked both conditions equally for clarity, PE higher than HT for accuracy and HT higher than PE for style. When asked to choose their favourite translation, HT obtained a higher percentage of preferences than PE: 63\% to 37\%.



Bowker~\shortcite{Bowker2009} presented French- and English-speaking minorities in Canada four translations (HT, full PE, light PE and raw MT) of three short governmental texts (approximately 325 words) written in a relatively clear, neutral style with reasonably short sentences.
Preference took into account not only the quality of the translations but also the time and cost required to produce them.
The results were rather different for each group.
In the French-speaking minority group 71\% preferred HT versus 29\% PE (21\% full and 8\% light). However, half of the participants were language professionals, which skewed the results. In fact, when they were removed the results changed drastically: 56\% preferred HT versus 44\% PE (29\% full, 15\% light). 
As for the English minority, 8\% preferred HT versus 87\% PE (38\% full, 49\% light) and 4\% raw MT.

Bowker and Buitrago Ciro~\shortcite{Bowker2015} presented Spanish-speaking immigrants in Canada with four translations (HT, full PE, light PE and raw MT) of three short texts (301 to 380 words) containing library-related information and asked them which they prefered.
PE and HT attained a similar number of preferences; 49\% of respondents preferred PE (24\% full and 25\% light, respectively) compared to 42\% who preferred HT.
Raw MT lagged considerably behind with the remaining 9\% of the preferences.

Green et al.~\shortcite{Green2013}~assessed the quality of HT versus PE for Wikipedia articles translated from English into Arabic, French and German. Quality was measured by means of preference, done by ranking on isolated sentences via crowdsourcing. PE was found significantly better for all translation directions.

\begin{table*}
\centering
\begin{tabular}{| l | l | l | l | r | l|}
\hline
\bf Dataset	& \bf Direction				& \bf PE type	& \bf MT systems		& \# \bf Sent. pairs	& \bf Domain\\
\hline
Taraxü	& en$\rightarrow$de	& Light\footnotemark			& \multirow{3}{*}{2 SMT, 2 RBMT}				& 272				& \multirow{3}{*}{News}\\
		& de$\rightarrow$en	& 			&	& 240&\\
							& es$\rightarrow$de\footnotemark	& 			&	& 101&\\
\hline
\multirow{2}{*}{IWSLT}		& en$\rightarrow$de		& \multirow{2}{*}{Light} & 4 NMT, 4 SMT				& \multirow{2}{*}{600}				& \multirow{2}{*}{Subtitles}\\
							& en$\rightarrow$fr		& 	& 2 NMT, 3 SMT				& 				&\\
\hline
MS 	& zh$\rightarrow$en		& Full			& 1 NMT\footnotemark			& 1,000\footnotemark			& News\\
\hline
\end{tabular}
\caption{Information about the datasets used in the experiments}
\label{t_datasets}
\end{table*}

\subsection{Characteristics of HT and PE}

Daems et al.~\shortcite{daems_17} used HT and PE news translated from English into Dutch. 
They presented them to translation students and colleagues, whose task was to identify which translations were PE.
They also tried an automated approach, for which they built a classifier with 55 features using surface forms and linguistic information at lexical, syntactic and semantic levels. 
No proof of the existence of post-editese was found, either perceived (students) or measurable (classifier).

Čulo and Nitzke~\shortcite{DBLP:conf/eamt/CuloN16} compared  MT, PE and HT in terms of terminology and found that the way terminology is used in PE is closer to MT than to HT and has less variation than in HT.

Farrell~\shortcite{farrell_tc40} identified MT markers (i.e. ``translation solutions which occurred with a statistically significantly higher frequency in PEMT than in HT'') in short texts from Wikipedia translated from English into Italian and found that MT tends to choose a subset of all the possible translation solutions (the most frequent ones) and that this is the case also, to some extent, in PEs.
HTs and PEs were also compared in terms of number of errors, which were found to be comparable, corroborating the findings of the literature covered at the beginning of this section.

Carl and Schaeffer~\shortcite{carl2017} studied a dataset with several HTs and PEs from different professional translators for the same source text and found out that PEMT, using a statistical MT system, leads to more literal translations, which also have less lexical variation.
Closely related to this and in a similar study, Bangalore et al.~\shortcite{Bangalore:572317} found PEs to have less syntactic variation than HTs.

Our contribution falls into this research line, to which we contribute a computational study whose analyses are chosen to align to translation universals and laws of translation and that covers multiple languages and domains.

\footnotetext[3]{``Translators were asked to perform only the minimal post-editing necessary to achieve an acceptable translation quality.'' \cite{avramidis2014Taraxu}}
\footnotetext[4]{This dataset contains an additional translation direction (de$\rightarrow$es) which is not used here due to its small size; 40 sentence pairs.}
\footnotetext[5]{The MT systems used in the MT and in the PE condition are not the same. The one in the MT condition is the best system in Hassan et al.~\shortcite{hassan2018parity} while the one in the PE condition is Google Translate, again as provided in Hassan et al.~\shortcite{hassan2018parity}.}
\footnotetext{The original dataset contains 2,001 sentences. We only use the subset in which the source text is original instead of translationese~\cite{toral-EtAl:2018:WMT}.}

\section{Experiments}\label{s_experiments}

In this section we first describe the datasets used (Section~\ref{s_datasets}), and then report on each of the experiments that we carried out in the subsequent subsections: lexical variety (Section~\ref{s_lexical_variety}), lexical density (Section~\ref{s_lexical_density}), length ratio (Section~\ref{s_length_ratio}) and part-of-speech sequences (Section~\ref{s_pp_diff}).

\subsection{Datasets}\label{s_datasets}
We make use of three datasets in all our experiments: Taraxü~\cite{avramidis2014Taraxu}, IWSLT~\cite{cettolo2015iwslt,cettolo2016iwslt} and Microsoft ``Human Parity''~\cite{hassan2018parity}, henceforth referred to as MS.
These datasets cover five different translation directions that involve five languages:\footnote{In the tables and experiments we will refer to languages with their ISO-2 codes.} English$\leftrightarrow$German, English$\rightarrow$French, Spanish$\rightarrow$German and Chinese$\rightarrow$English.
In addition, this choice of datasets allows us to include a longitudinal aspect into the analyses since there are state-of-the-art MT systems from almost one decade ago (in Taraxü), from three and four years ago (IWSLT) and from just one year ago (MS).
Table~\ref{t_datasets} shows detailed information about each dataset, namely its translation direction(s), type of PE done, paradigm of the MT system(s) used, number of sentence pairs and domain of its text.


\begin{table*}
\centering
\begin{tabular}{ |l|r|r|r|r| r|r|}
\hline
\bf Translation	& \multicolumn{6}{c|}{\bf Dataset and translation direction}\\
\cline{2-7}
\bf type		& \multicolumn{3}{ c| }{Taraxü} & \multicolumn{2}{ c| }{IWSLT}& \multicolumn{1}{ c| }{MS}  \\
\cline{2-7}
				&de$\rightarrow$en & en$\rightarrow$de & es$\rightarrow$de & en$\rightarrow$de & en$\rightarrow$fr & zh$\rightarrow$en\\
\hline
HT				& 0.26 & 0.27 & 0.31 & 0.20 & 0.16 & 0.14\\
\hline
PE				& {\bf -2.05\%}	& {\bf -1.81\%}	& {$\dagger$\bf -1.27\%}	&{\bf -3.86\%}	&{\bf -1.17\%} & {\bf -4.76\%}\\
MT 				&-2.94\%& -3.62\%	&-5.91\%	&-10.93\%	&-6.04\%&-6.96\%\\
\hline
PE-NMT		&			&			&			&-4.21\%	&-1.88\%&-4.76\%\\
PE-SMT		&{\bf -1.59\%}	& {\bf -1.31\%}&{$\dagger$\bf -1.03\%}	&{\bf -3.50\%}	&{\bf -0.70\%}&\\

PE-RBMT		&-2.79\%	& -2.04\%&-3.05\%	&			&			&\\
\hline
NMT			&			&			&			&-12.22\%	&-8.18\%	&-7.33\%\\
SMT			&{\bf -2.36\%}	& {\bf -2.36\%}&{\bf -6.42\%}	&{\bf -9.63\%}	&{\bf -4.61\%}	&\\
RBMT		&-3.08\%& -4.26\%	&-7.78\%	&			&			&\\
\hline
\end{tabular}
\caption{TTR scores for HT and relative differences for PE and MT. For directions with more than one MT system, the result shown in rows PE and MT uses the average score of all the PEs or MT outputs, respectively. The best result (highest TTR) in each group of rows is shown in bold. If a $\dagger$ is not shown then the TTR for HT is significantly higher than the TTRs for all the translations in that cell (the 95\% confidence interval of the TTR of HT, obtained with bootstrap resampling, is higher and there is no overlap).}
\label{t_ttr}
\end{table*}

We note the following two limitations in some of the datasets:
\begin{itemize}
\item Mismatch of translator competence. Both PE and HT are carried out by professional translators in two of the datasets (Taraxü and MS). 
However, in the remaining one, IWSLT, professional translators do PE, while the translators doing HT are not necessarily professionals\footnote{``We accept all fluently bilingual volunteers as translators'', \url{https://translations.ted.com/TED_Translator_Resources:_Main_guide#Translation}}.
Thus, if we find differences between PEs and HTs, for this dataset this may not be (entirely) due to the two translations procedures leading to different translations but (also) to the different translations being produced by translators with different levels of proficiency.
\item Source language being translationese. In two of the datasets (MS and IWSLT), the source language and the language in which those texts were originally written is the same. This is not the case however for Taraxü, for which the original language of the source texts is Czech. 
We can still compare MT to PE although we need to take into account that these texts are easier for MT than original texts~\cite{toral-EtAl:2018:WMT}. However the comparison between PE (or MT) and HT is problematic since the HT was not translated from the source language but from another language (Czech).
\end{itemize}

\subsection{Lexical Variety}\label{s_lexical_variety}

We assess the lexical variety of a translation (MT, PE or HT) by calculating its type-token ratio (TTR), as shown in equation \ref{eq_ttr}.

\begin{equation}\label{eq_ttr}
TTR = \frac{number~of~types}{number~of~tokens}
\end{equation}

Farrell~\shortcite{farrell_tc40} observed that MT tends to produce a subset of all the possible translations in the target language (the ones used most frequently in the training data).
Therefore, we hypothesise the TTR of MT, and by extension that of PE too, to be lower than that of HT.
If this is the case, then PE would be, in terms of lexical variety, simpler than HT.

Table~\ref{t_ttr} shows the results for each dataset and language direction.
In all cases the lexical variety in PE is lower than in HT, and again in all cases, that of MT is lower than that of PE.
This could be interpreted as follows: 
(i) lexical variety is low in MT because these systems prefer the translation solutions most frequently used in the training data; (ii) a post-editor will add lexical variety to some degree, but because MT primes him/her, the resulting translation will not achieve the level of lexical diversity that is attained in HT.

We now look at the results of PE and MT for different MT paradigms.
In the Taraxü dataset  we can compare rule-based and statistical MT systems. Rule-based MT has a lower TTR in all three translation directions of this dataset and this is then reflected in a lower TTR again when a system of this paradigm is used for post-editing.
In the IWSLT dataset we can confront statistical and neural MT systems. In all cases the lexical variety of neural MT is lower than that of statistical MT. Again, the same trend shows when we look at their PEs.
This is perhaps a surprising result, since NMT systems outperformed SMT in the IWSLT dataset, in terms of HTER~\cite{cettolo2015iwslt}.

\begin{table*}
\centering
\begin{tabular}{ |l|r|r|r|r| r|r|}
\hline
\bf Translation	& \multicolumn{6}{c|}{\bf Dataset and translation direction}\\
\cline{2-7}
\bf Type		& \multicolumn{3}{ c|}{Taraxü} & \multicolumn{2}{c|}{IWSLT} & \multicolumn{1}{ c| }{MS}\\
\cline{2-7}
				&	de$\rightarrow$en & en$\rightarrow$de & es$\rightarrow$de & en$\rightarrow$de & en$\rightarrow$fr & zh$\rightarrow$en\\
\hline
HT				&0.55 & 0.53 & 0.53 & 0.48 & 0.46 & 0.59\\
\hline
PE				&-1.00\% & -2.48\% & \bf{-4.31\%} & \bf{-3.46\%} & -1.24\% & \bf{-0.46\%}\\
MT 				&\bf{-0.81\%} & \bf{-0.69\%} & -4.53\% & -5.14\% & \bf{-0.94\%} & -2.37\%\\
\hline
PE-NMT		&			&		&			& -3.88\%  & -1.47\% & -0.46\%\\
PE-SMT		&\bf{-0.54\%} & -2.87\% & -4.78\% & \bf{-3.04\%} & \bf{-1.09\%} & \\
PE-RBMT		& -1.46\% & \bf{-2.09\%} & \bf{-3.84\%} &			&		 	&\\
\hline
NMT			&			&			&		   & -6.31\%   & -3.14\%   & -2.37\%\\
SMT			& \bf{-0.80\%}   & \bf{0.14\%}  & \bf{-3.45\%} & \bf{-3.98\%}   & \bf{0.53\%}		&\\\
RBMT			& -0.83\%   & -1.51\% & -5.61\% &				&		&\\
\hline
\end{tabular}
\caption{Lexical density scores for HT and relative differences for PE and MT. For directions with more than one MT system, the result shown in rows PE and MT uses the average score of all the PEs or MT outputs, respectively. The best result (highest density) in each group of rows is shown in bold.}
\label{t_lex_density}
\end{table*}

\subsection{Lexical Density}\label{s_lexical_density}

Lexical density measures the amount of information present in a text by calculating the ratio between the number of content words (adverbs, adjectives, nouns and verbs) and its total number of words, as shown in equation~\ref{eq_lex_density}.

\begin{equation}\label{eq_lex_density}
lex\_density = \frac{number~of~content~words}{number~of~total~words}
\end{equation}

Translationese has been found to have a lower percentage of content words than original texts, thus being, from this point of view, lexically simpler~\cite{scarpa2006corpus}.
To identify and count content words we tag the target sides of the datasets with their parts-of-speech (PoS) using UDPipe~\cite{straka2016udpipe}, a PoS tagger that uses the Universal PoS tagset.\footnote{\url{https://universaldependencies.org/u/pos/}}
Then we assess the lexical density of each translation (HT, PE and MT) using this PoS-tagged version\footnote{UDPipe's PoS tagging F1-score is over 90\% for all the three target languages considered: de, en and fr~\cite[Table~2]{udpipe:2017} } of the datasets.

Table~\ref{t_lex_density} shows the results.
In both PE and MT the lexical density is lower than in HT. However between PE and MT, there is no systematic distinction.
When inspecting PEs using different MT paradigms, we do not find any clear trend between SMT and RBMT, but one such trend shows up when we inspect SMT and NMT: in the two comparisons we can establish in our dataset (the two translation directions in the IWSLT dataset), PE-NMT leads to a lower lexical density than PE-SMT.
Finally, looking at MT outputs produced by different types of MT systems, we observe that both RBMT and NMT lead to lower lexical densities than SMT.

%

\subsection{Length Ratio}\label{s_length_ratio}

Given a source text $ST$ and a target text $TT$, i.e. $TT$ is a translation of the $ST$ (HT, PE or MT), we compute the absolute difference in length (measured in characters) between the two, normalised by the length of the $ST$, as shown in equation \ref{eq_len}.

\begin{equation}\label{eq_len}
length~ratio = \frac{| length_{ST} - length_{TT} |}{length_{ST}}
\end{equation}


Because (i) MT results in a translation of similar length to that of the ST,\footnote{This is necessary the case for RBMT and SMT as the number of TL tokens they can produce per each SL token is limited; e.g. the longest a translation with SMT can be is the number of tokens in the ST multiplied by the longest phrase in the phrase table, which is typically 7. NMT does not have this limitation, so we do not argue in this direction for that MT paradigm.}
and PE is primed by the MT output while (ii) a translator working from scratch (HT) may translate more freely in terms of length,
we hypothesise that the difference in absolute length is smaller for MT and PE than it is for HT.
If this is true, it would be a case of interference in PE, as the typical length of sentences translated with this method would be similar to the length used in the source text.


\begin{table}[htbp]
\centering
\begin{tabular}{ |l|l|r|r|r|}
\hline
\bf \multirow{2}{*}{Dataset}	& \bf \multirow{2}{*}{Direct.} & \multicolumn{3}{c|}{\bf Length ratio}\\
\cline{3-5}
										&						& \bf HT	& \bf PE	& \bf MT\\
\hline
\multirow{3}{*}{Taraxü}		& de$\rightarrow$en	& 0.16		& $\ddagger$-38.5\%		& $\dagger$-36.9\%\\
									& en$\rightarrow$de	& 0.22	& $\ddagger$-33.4\%		& $\ddagger$-38.5\%\\
									& es$\rightarrow$de	& 0.17	& *-25.2\%		& -21.0\%\\
\hline
\multirow{2}{*}{IWSLT}		& en$\rightarrow$de	& 0.17	& -3.4\%		& $\dagger$-18.8\%\\
									& en$\rightarrow$fr		& 0.18	& 6.7\%		& -10.9\%\\
\hline
MS 								& zh$\rightarrow$en	& 1.41	& $\ddagger$-9.9\%		& $\ddagger$-9.1\%\\
\hline
\end{tabular}
\caption{Length ratio scores for HT and relative differences for PE and MT. For directions with more than one MT system, the result shown in columns PE and MT uses the average length ratio of all the PEs or MT outputs, respectively. * indicates that the score for HT is significantly higher with $\alpha=0.05$ ($\dagger$ with $\alpha=0.01$ and $\ddagger$ with $\alpha=0.001$) than the scores of all the PEs/MTs represented in the cell, based on one-tailed paired t-tests.}
\label{t_lenratio}
\end{table}

We compute this ratio at sentence level and average over all the sentences of the dataset.
Table \ref{t_lenratio} shows the results for each dataset and language direction.
The results in datasets Taraxü and MS match our hypothesis; in both datasets the length ratio is lower for PE and MT than it is for HT.
This is also the case for MT in dataset IWSLT.
However, in the results for PE in dataset IWSLT, the ratio of PE is actually higher than that of HT for en$\rightarrow$fr,  which seems to contradict our hypothesis. This may be attributed to the difference in translation proficiency between the translators that did HT and those that did PE that we commented upon in Section~\ref{s_datasets}. The latter are professional, while the first could be non-professional. It is known that non-professional translators tend to produce more literal translations, whose length should then be similar to that of the source text.

\begin{figure}
\includegraphics[width=0.49\textwidth]{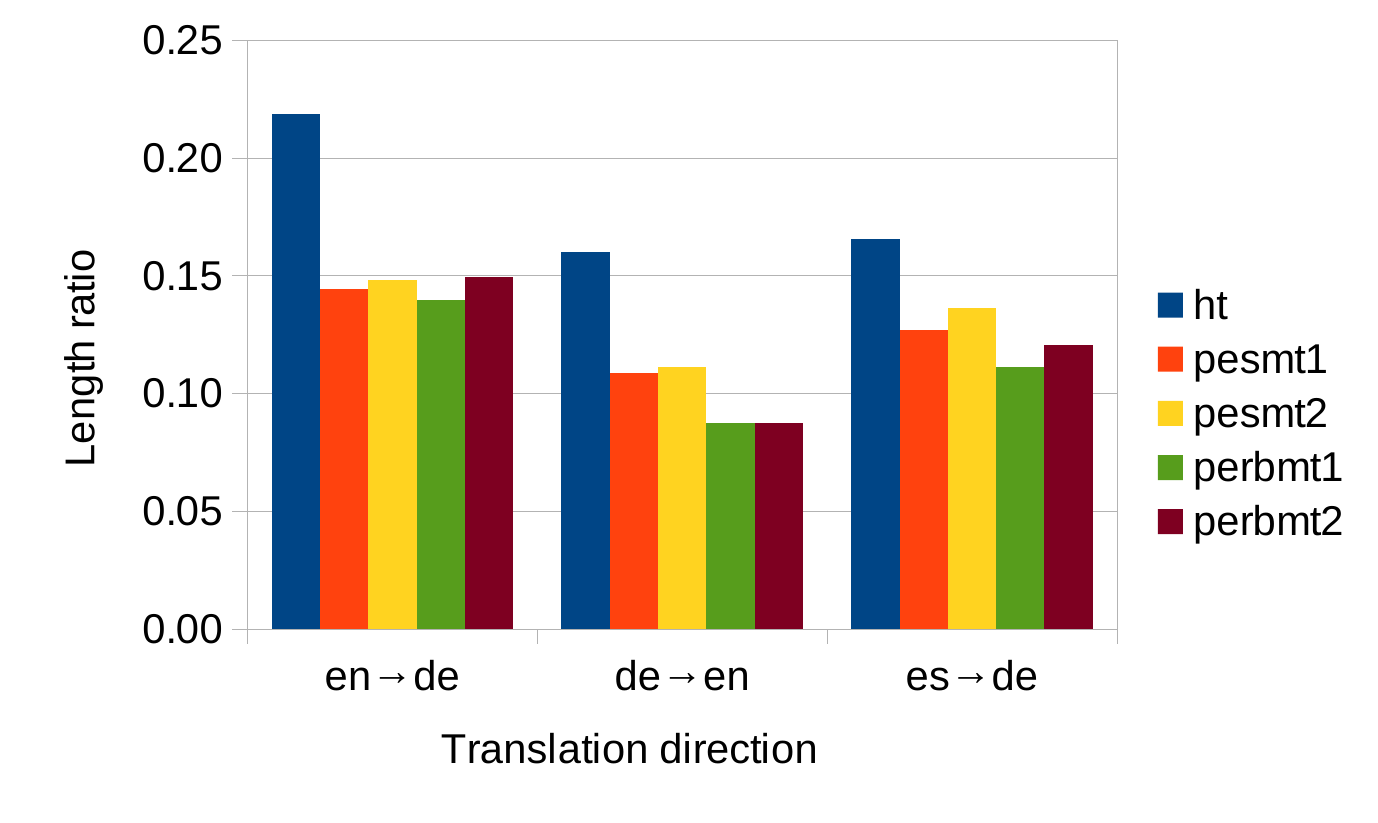}
\caption{Length ratio for HT and PEs in the Taraxü dataset.}
\label{f_lenratio_trx}
\end{figure}

We now look at the length ratio of PEs that use different MT systems.
Figure \ref{f_lenratio_trx} shows the length ratios of HTs and PEs that use different MT paradigms (SMT and RBMT) in the Taraxü dataset.
While for one of the translation direction (en$\rightarrow$de) the ratio of PE-SMT and PE-RBMT are similar, the two PE-RBMT systems have lower length ratios than the two PE-SMT systems in the other two translation directions.

\begin{table*}
\centering
\begin{tabular}{ |l|r|r|r|r| r|r|}
\hline
\bf Translation	& \multicolumn{6}{c|}{\bf Dataset and translation direction}\\
\cline{2-7}
\bf Type		& \multicolumn{3}{ c|}{Taraxü} & \multicolumn{2}{c|}{IWSLT} & \multicolumn{1}{ c| }{MS}\\
\cline{2-7}
				&	de$\rightarrow$en & en$\rightarrow$de & es$\rightarrow$de & en$\rightarrow$de & en$\rightarrow$fr & zh$\rightarrow$en\\
\hline
HT				&5.12 & 5.09 & 9.41 & 5.01 & 2.47 & 17.23\\
\hline
PE				&\bf{-13.84\%}	& \bf{-11.29\%}	& \bf{-8.58\%}	& \bf{-6.26\%}	& \bf{-2.03\%}	& -3.26\%\\
MT 				&-33.65\%	& -32.25\% & -20.71\% &-18.66\% & -11.07\% &\bf{-3.1\%}\\
\hline
PE-NMT		&	&	&	& \bf{-3.41\%} & \bf{-1.40\%}	& -3.26\%\\
PE-SMT		& \bf{-11.72\%} & -13.37\% & -10.48\% & -9.10\% & -2.46\% & \\
PE-RBMT		& -15.95\% & \bf{-9.20\%}  & \bf{-6.68\%} & & &\\
\hline
NMT			& 			&			&			& \bf{-5.89\%}    & \bf{-2.58\%} & -3.10\%\\
SMT			&\bf{-30.07\%} & -41.71\% & -26.30\% & -31.43\% & -7.95\% &\\
RBMT			&-37.24\% & \bf{-22.80\%} & \bf{-15.13\%} & &&\\
\hline
\end{tabular}
\caption{Perplexity difference scores for HT and relative differences for PE and MT. For directions with more than one MT system, the result shown in rows PE and MT uses the average score of all the PEs or MT outputs, respectively. The best result (highest perplexity) in each group of rows is shown in bold.}
\label{t_pp_diff}
\end{table*}

\subsection{Part-of-Speech Sequences}\label{s_pp_diff}

We assess the interference of the source language on a translation (HT, PE and MT) by measuring how close the sequence of PoS tags in the translation is to the typical PoS sequences of the source language and to the typical PoS sequences of the target language.
If the sequences of PoS tags used in a translation $A$ are more similar to the typical sequences of the source language than the sequences of another translation $B$, that could be an indication that $A$ has more interference from the source language than $B$.

Given a PoS-tagged translation $T$, a language model of a PoS-tagged corpus in the source language $LM_{SL}$ and a language model of a PoS-tagged corpus in the target language $LM_{TL}$, we calculate the difference of the perplexities of $T$ with respect to both language models, as shown in equation~\ref{eq_ppdiff}.

\begin{equation}\label{eq_ppdiff}
PP\_diff = PP(T, LM_{SL}) - PP(T, LM_{TL})
\end{equation}

A high result for a translation $T$ would indicate that $T$ is dissimilar to the source language (high perplexity with respect to the source language) and similar to the target language (low perplexity with respect to the target language). Conversely, a low result would indicate that $T$ is similar to the source language (low perplexity with respect to the source language) and dissimilar to the target language (high perplexity with respect to the target language).

Because MT systems are known to perform less reordering than human translators~\cite{toral-sanchez-cartagena-2017-multifaceted}, our hypothesis is that MT outputs, and by extension PEs, are more similar in terms of PoS sequences to the source language than HTs are.
This would mean that PEs and MT outputs have more interference in terms of PoS sequences than HTs.

For each language in our datasets (de, en, es, fr and zh), we PoS tag a
monolingual corpus\footnote{The corpora used for the different languages belong to the same domain, news. We use corpora of roughly the same size (around 100MB of text). This leads to corpora of between 1,617,527 sentences (es) and 2,187,421 (de).} with UDPipe, the PoS tagger already introduced in Section~\ref{s_lexical_density}.\footnote{While in Section~\ref{s_lexical_density} we PoS-tagged the target side of the datasets, in this experiment we PoS-tag both the source and target sides.
UDPipe's PoS tagging F1-score is over 90\% for four of the five languages involved (de, en, es and fr) and 83\% for the remaining one: zh~\cite[Table~2]{udpipe:2017}.
Given the lower performance of PoS tagging for zh, the results involving this language should be taken with caution.
}

We then build language models on these PoS tagged data with SRILM~\cite{stolcke2002srilm}, considering $n$-grams up to $n=6$, using interpolation and Witten-Bell smoothing.\footnote{The more advanced smoothing method Kneser-Ney did not work because the count-of-count statistics in our datasets are not suitable for this smoothing method, which may be due to the very small size of our vocabulary: the Universal Dependencies PoS tagset.}
Because we use the Universal PoS tagset, the set of PoS tags is the same for all the languages, which means that all our language models share the same vocabulary.

Table~\ref{t_pp_diff} shows the results.
In terms of HT versus PE and MT, we observe similar trends to those observed earlier for lexical variety (Table~\ref{t_ttr}), namely MT obtains the lowest perplexity difference score and HT the highest, with PE lying somewhere between the two.
The only exception to this is seen in the MS dataset, where the value for MT is slightly higher than that for PE (-3.1\% versus -3.26\%). It should be taken into account, as already explained in Section~\ref{s_datasets}, that the MT systems in the MT and PE conditions are different in this dataset, with the one in the MT condition being substantially better~\cite{hassan2018parity}.
Overall, we interpret these results as MT being the translation type that contains the most interference in terms of PoS sequences, followed by PE.

We now look at the PE and MT results under different MT paradigms.
Comparing SMT and NMT, the results indicate that the latter has less interference, both in PE and MT conditions. This corroborates earlier research that compared SMT and NMT in terms of reordering~\cite{DBLP:conf/emnlp/BentivogliBCF16}.
We do not find clear trends when comparing SMT and RBMT though.

%
%
%
%
%
%
%
%

%
%
%
%
%
%
%

\section{Conclusions and Future Work}\label{s_conclusions}

We have carried out a set of computational analyses on three datasets that contain five translation directions with the aim of finding out whether post-edited translations (PEs) exhibit different phenomena than human translations from scratch (HTs). In other words, whether there is evidence of the existence of {\it post-editese}.
The analyses conducted measure aspects related to translation universals and laws of translation, namely simplification, normalisation and interference.
With these analysis, we find evidence of post-editese (RQ1), whose main characteristics (RQ2) we summarise as follows:

\begin{itemize}
\item PEs have lower lexical variety and lower lexical density than HTs. We link these to the simplification principle of translationese. Thus, these results indicate that {\it post-editese} is lexically simpler than translationese.
\item Sentence length in PEs is more similar to the sentence length of the source texts, than sentence length in HTs. We link this finding to interference and normalisation: (i) PEs have interference from the source text in terms of length, which leads to translations that follow the typical sentence length of the source language; (ii) this results in a target text whose length tends to become normalised.
\item Part-of-speech (PoS) sequences in PEs are more similar to the typical PoS sequences of the source language, than PoS sequences in HTs.
We link this to the interference principle: the sequences of grammatical units in PEs preserve to some extent the sequences that are typical of the source language.
\end{itemize}

In the paper we have not considered only HTs and PEs but also MT outputs, from the MT systems that were the starting point to produce the PEs.
This to corroborate a claim in the literature~\cite{Green2013}, namely that in PE the translator is primed by the MT output.
We expected then to find similar trends to those found in PEs also in MT outputs and this was indeed the case in all four experiments.
In two of the experiments, lexical variety and PoS sequences, the results of PE were somewhere in the middle between those of HT and MT.
Our interpretation is that a post-editor improves the initial MT output in terms of variety and PoS sequences, but due to being primed by the MT output, the result cannot attain the level of HT, and the footprint of the MT system remains in the resulting PE.

We have also looked at different MT paradigms (rule-based, statistical and neural), to find out whether these lead to different characteristics in the resulting PEs (RQ3).
Neural MT diminishes to some extent the interference in terms of PoS sequences, probably because it is better at reordering~\cite{DBLP:conf/emnlp/BentivogliBCF16}.
Statistical MT has obtained better results than the other two MT paradigms in terms of lexical variety, and also better than neural MT, but on par with rule-based MT, in lexical density.

In a nutshell, we have found that PEs tend to be simpler and more normalised and to have a higher degree of interference from the source text than HTs.
This seems to be caused because these characteristics are already present in the MT outputs that are the starting point of the PEs.
We find thus evidence of {\it post-editese}, which can be thought of as an exacerbated translationese.

While PE is very useful in terms of productivity, which arguably is the main reason behind its wide adoption in industry, the findings of this paper flag a potential issue.
Because PEs are simpler and have a higher interference from the source language than HTs, the extensive use of PE rather than HT may have serious implications for the target language in the long term, for example that it becomes impoverished (simplification) and overly influenced by the source language (interference).
At the same time, we have shown that these issues  cannot be attributed to PE {\it per se} but that they originate in the MT systems used as the starting point for PE. Identifying these issues might be then the first step for further research on addressing these problems in current state-of-the-art MT systems.

Throughout the paper, we have assumed that lexical diversity and density correlate directly with translation quality; i.e. the more diverse and dense a translation the better.
In this regard, we acknowledge that in translation there is a tension between diversity and consistency, especially in technical translation.
At the same time, none of our datasets falls under the domain of technical texts.

We also acknowledge that our study is based on rather superficial linguistic features, either at surface or morphological level (PoS tags).
For future work, therefore, we plan to explore the use of additional features, especially relying on deeper linguistic analyses.
In addition, we plan to study the overlap between multiple HTs and PEs for the same text, to assess whether it is higher between PEs, which would indicate a higher degree of homogeneisation.

Another line we would like to pursue is that of automatic discrimination between PE and HT. While this has been shown to be possible with a high degree of accuracy between original texts and HTs, this is not the case for PE versus HT in the attempts conducted to date~\cite{daems_17}.

Finally, we would like to point out that all our code and data are publicly released,\footnote{\url{https://github.com/antot/posteditese_mtsummit19}} so we encourage interested parties
to use these resources to conduct further analyses.




\section*{Acknowledgments}

I would like to thank Luisa Bentivogli, Joke Daems, Michael Farrell, Lieve Macken and Maja Popović for pointing me to relevant datasets and related papers and for insightful discussions on the topic of this paper.
I would also like to thank the reviewers; their comments have definitely led to improve this paper.

\bibliography{mtsummit2019}

\begin{thebibliography}{}

\bibitem[\protect\citename{Avramidis \bgroup et al.\egroup
  }2014]{avramidis2014Taraxu}
Avramidis, Eleftherios, Aljoscha Burchardt, Sabine Hunsicker, Maja Popovic,
  Cindy Tscherwinka, David Vilar, and Hans Uszkoreit.
\newblock 2014.
\newblock The tarax{\"u} corpus of human-annotated machine translations.
\newblock In {\em LREC}, pages 2679--2682.

\bibitem[\protect\citename{Baker}1993]{baker1993corpus}
Baker, Mona.
\newblock 1993.
\newblock Corpus linguistics and translation studies: Implications and
  applications.
\newblock {\em Text and technology: In honour of John Sinclair}, 233:250.

\bibitem[\protect\citename{Bangalore \bgroup et al.\egroup
  }2015]{Bangalore:572317}
Bangalore, Srinivas, Bergljot Behrens, Michael Carl, Maheshwar Gankhot, Arndt
  Heilmann, Jean Nitzke, Moritz Schaeffer, and Annegret Sturm.
\newblock 2015.
\newblock {T}he role of syntactic variation in translation and post-editing.
\newblock {\em Translation spaces}, 4(1):119--144.

\bibitem[\protect\citename{Bentivogli \bgroup et al.\egroup
  }2016]{DBLP:conf/emnlp/BentivogliBCF16}
Bentivogli, Luisa, Arianna Bisazza, Mauro Cettolo, and Marcello Federico.
\newblock 2016.
\newblock Neural versus phrase-based machine translation quality: a case study.
\newblock In Su, Jian, Xavier Carreras, and Kevin Duh, editors, {\em
  Proceedings of the 2016 Conference on Empirical Methods in Natural Language
  Processing, {EMNLP} 2016, Austin, Texas, USA, November 1-4, 2016}, pages
  257--267. The Association for Computational Linguistics.

\bibitem[\protect\citename{Bowker and {Buitrago Ciro}}2015]{Bowker2015}
Bowker, Lynne and Jairo {Buitrago Ciro}.
\newblock 2015.
\newblock {Investigating the usefulness of machine translation for newcomers at
  the public library}.
\newblock {\em Translation and Interpreting Studies}, 10(2):165--186.

\bibitem[\protect\citename{Bowker}2009]{Bowker2009}
Bowker, Lynne.
\newblock 2009.
\newblock Can machine translation meet the needs of official language minority
  communities in canada? a recipient evaluation.
\newblock {\em Linguistica Antverpiensia, New Series--Themes in Translation
  Studies}, (8).

\bibitem[\protect\citename{Carl and Schaeffer}2017]{carl2017}
Carl, Michael and {Moritz Jonas} Schaeffer.
\newblock 2017.
\newblock Why translation is difficult: A corpus-based study of non-literality
  in post-editing and from-scratch translation.
\newblock {\em Hermes}, 56:43--57.

\bibitem[\protect\citename{Cettolo \bgroup et al.\egroup
  }2015]{cettolo2015iwslt}
Cettolo, Mauro, Jan Niehues, Sebastian St{\"u}ker, Luisa Bentivogli, Roldano
  Cattoni, and Marcello Federico.
\newblock 2015.
\newblock The iwslt 2015 evaluation campaign.
\newblock In {\em IWSLT 2015, International Workshop on Spoken Language
  Translation}.

\bibitem[\protect\citename{Daems \bgroup et al.\egroup }2017]{daems_17}
Daems, Joke, Orph{\'e}e De~Clercq, and Lieve Macken.
\newblock 2017.
\newblock Translationese and post-editese : how comparable is comparable
  quality?
\newblock {\em LINGUISTICA ANTVERPIENSIA NEW SERIES-THEMES IN TRANSLATION
  STUDIES}, 16:89--103.

\bibitem[\protect\citename{Farrell}2018]{farrell_tc40}
Farrell, Michael.
\newblock 2018.
\newblock {Machine Translation Markers in Post-Edited Machine Translation
  Output}.
\newblock In {\em Proceedings of the 40th Conference Translating and the
  Computer}, pages 50--59.

\bibitem[\protect\citename{Fiederer and O’Brien}2009]{fiederer2009quality}
Fiederer, Rebecca and Sharon O’Brien.
\newblock 2009.
\newblock Quality and machine translation: A realistic objective.
\newblock {\em The Journal of Specialised Translation}, 11:52--74.

\bibitem[\protect\citename{Garcia}2010]{garcia2010}
Garcia, Ignacio.
\newblock 2010.
\newblock Is machine translation ready yet?
\newblock {\em Target. International Journal of Translation Studies},
  22(1):7--21.

\bibitem[\protect\citename{Green \bgroup et al.\egroup }2013]{Green2013}
Green, Spence, Jeffrey Heer, and Christopher~D Manning.
\newblock 2013.
\newblock {The efficacy of human post-editing for language translation}.
\newblock {\em Chi 2013}, pages 439--448.

\bibitem[\protect\citename{Guerberof}2009]{guerberof2009productivity}
Guerberof, Ana.
\newblock 2009.
\newblock Productivity and quality in mt post-editing.
\newblock In {\em MT Summit XII-Workshop: Beyond Translation Memories: New
  Tools for Translators MT}. Citeseer.

\bibitem[\protect\citename{Hassan \bgroup et al.\egroup
  }2018]{hassan2018parity}
Hassan, Hany, Anthony Aue, Chang Chen, Vishal Chowdhary, Jonathan Clark,
  Christian Federmann, Xuedong Huang, Marcin Junczys-Dowmunt, Will Lewis,
  Mu~Li, Shujie Liu, Tie-Yan Liu, Renqian Luo, Arul Menezes, Tao Qin, Frank
  Seide, Xu~Tan, Fei Tian, Lijun Wu, Shuangzhi Wu, Yingce Xia, Dongdong Zhang,
  Zhirui Zhang, and Ming Zhou.
\newblock 2018.
\newblock {Achieving Human Parity on Automatic Chinese to English News
  Translation}.
\newblock \url{https://arxiv.org/abs/1803.05567}.

\bibitem[\protect\citename{Koponen}2016]{Koponen2016}
Koponen, Maarit.
\newblock 2016.
\newblock {Is machine translation post-editing worth the effort? A survey of
  research into post-editing and effort}.
\newblock {\em Journal of Specialised Translation}, 25(25):131--148.

\bibitem[\protect\citename{Mauro \bgroup et al.\egroup }2016]{cettolo2016iwslt}
Mauro, Cettolo, Niehues Jan, St{\"u}ker Sebastian, Bentivogli Luisa, Cattoni
  Roldano, and Federico Marcello.
\newblock 2016.
\newblock The iwslt 2016 evaluation campaign.
\newblock In {\em International Workshop on Spoken Language Translation}.

\bibitem[\protect\citename{Plitt and Masselot}2010]{plitt2010productivity}
Plitt, Mirko and Fran{\c{c}}ois Masselot.
\newblock 2010.
\newblock A productivity test of statistical machine translation post-editing
  in a typical localisation context.
\newblock {\em The Prague bulletin of mathematical linguistics}, 93:7--16.

\bibitem[\protect\citename{Scarpa}2006]{scarpa2006corpus}
Scarpa, Federica.
\newblock 2006.
\newblock Corpus-based quality-assessment of specialist translation: A study
  using parallel and comparable corpora in english and italian.
\newblock {\em Insights into specialized translation--linguistics insights.
  Bern: Peter Lang}, pages 155--172.

\bibitem[\protect\citename{Stolcke}2002]{stolcke2002srilm}
Stolcke, Andreas.
\newblock 2002.
\newblock Srilm-an extensible language modeling toolkit.
\newblock In {\em Seventh international conference on spoken language
  processing}.

\bibitem[\protect\citename{Straka and Strakov\'{a}}2017]{udpipe:2017}
Straka, Milan and Jana Strakov\'{a}.
\newblock 2017.
\newblock Tokenizing, pos tagging, lemmatizing and parsing ud 2.0 with udpipe.
\newblock In {\em Proceedings of the CoNLL 2017 Shared Task: Multilingual
  Parsing from Raw Text to Universal Dependencies}, pages 88--99, Vancouver,
  Canada, August. Association for Computational Linguistics.

\bibitem[\protect\citename{Straka \bgroup et al.\egroup
  }2016]{straka2016udpipe}
Straka, Milan, Jan Hajic, and Jana Strakov{\'a}.
\newblock 2016.
\newblock Udpipe: Trainable pipeline for processing conll-u files performing
  tokenization, morphological analysis, pos tagging and parsing.
\newblock In {\em LREC}.

\bibitem[\protect\citename{Toral and
  S{\'a}nchez-Cartagena}2017]{toral-sanchez-cartagena-2017-multifaceted}
Toral, Antonio and V{\'\i}ctor~M. S{\'a}nchez-Cartagena.
\newblock 2017.
\newblock A multifaceted evaluation of neural versus phrase-based machine
  translation for 9 language directions.
\newblock In {\em Proceedings of the 15th Conference of the European Chapter of
  the Association for Computational Linguistics: Volume 1, Long Papers}, pages
  1063--1073, Valencia, Spain, April. Association for Computational
  Linguistics.

\bibitem[\protect\citename{Toral \bgroup et al.\egroup
  }2018]{toral-EtAl:2018:WMT}
Toral, Antonio, Sheila Castilho, Ke~Hu, and Andy Way.
\newblock 2018.
\newblock Attaining the unattainable? reassessing claims of human parity in
  neural machine translation.
\newblock In {\em Proceedings of the Third Conference on Machine Translation,
  Volume 1: Research Papers}, pages 113--123, Belgium, Brussels, October.
  Association for Computational Linguistics.

\bibitem[\protect\citename{Toury}2012]{toury2012descriptive}
Toury, Gideon.
\newblock 2012.
\newblock {\em Descriptive translation studies and beyond: Revised edition},
  volume 100.
\newblock John Benjamins Publishing.

\bibitem[\protect\citename{Čulo and Nitzke}2016]{DBLP:conf/eamt/CuloN16}
Čulo, Oliver and Jean Nitzke.
\newblock 2016.
\newblock Patterns of terminological variation in post-editing and of cognate
  use in machine translation in contrast to human translation.
\newblock In {\em Proceedings of the 19th Annual Conference of the European
  Association for Machine Translation, {EAMT} 2017, Riga, Latvia, May 30 - June
  1, 2016}, pages 106--114. European Association for Machine Translation.

\end{thebibliography}
\bibliographystyle{mtsummit2019}

\end{document}